\PassOptionsToPackage{most}{tcolorbox}

\documentclass[11pt]{article}

\usepackage[final]{acl}
\usepackage{comment}
\usepackage{amsmath}
\usepackage{times}
\usepackage{latexsym}
\usepackage{xcolor}
\usepackage{tcolorbox}
\tcbuselibrary{skins,breakable}
\usepackage{booktabs}
\usepackage[T1]{fontenc}
\usepackage{enumitem}

\usepackage[utf8]{inputenc}

\usepackage{microtype}

\usepackage{inconsolata}

\usepackage{graphicx}

\title{Instructions for *ACL Proceedings}

\author{Shidong Cao$^1$, Hongzhan Lin$^1$, Yuxuan Gu$^2$, Ziyang Luo$^1$, \textbf{Jing Ma}$^{1}$\thanks{Corresponding author.} \\ 
        $^1$Hong Kong Baptist University\\
        $^2$Harbin Institute of Technology\\
        \texttt{\{cssdcao,cshzlin,majing\}@comp.hkbu.edu.hk}}
\title{\textsc{DiffCoT}: Diffusion-styled Chain-of-Thought Reasoning in LLMs}

\begin{document}
\maketitle
\begin{abstract}
Chain-of-Thought (CoT) reasoning improves multi-step mathematical problem solving in large language models but remains vulnerable to exposure bias and error accumulation, as early mistakes propagate irreversibly through autoregressive decoding. In this work, we propose \textsc{DiffCoT}, a diffusion-styled CoT framework that reformulates CoT reasoning as an iterative denoising process. \textsc{DiffCoT} integrates diffusion principles at the reasoning-step level via a sliding-window mechanism, enabling unified generation and retrospective correction of intermediate steps while preserving token-level autoregression. To maintain causal consistency, we further introduce a causal diffusion noise schedule that respects the temporal structure of reasoning chains. Extensive experiments on three multi-step CoT reasoning benchmarks across diverse model backbones demonstrate that \textsc{DiffCoT} consistently outperforms existing CoT preference optimization methods, yielding improved robustness and error-correction capability in CoT reasoning.
\end{abstract}

\section{Introduction}

Large Language Models (LLMs)~\citep{brown2020language,achiam2023gpt,guo2025deepseek} have marked a major breakthrough in natural language processing, achieving competitive performance across a wide range of mathematical reasoning tasks. A widely adopted technique in LLMs, Chain-of-Thought (CoT) reasoning~\citep{wei2022chain,kojima2022large}, enhances mathematical problem-solving by decomposing tasks into step-by-step intermediate reasoning. However, CoT reasoning is highly sensitive to potential errors, where mistakes introduced in the early stages can propagate through later steps, often leading to incorrect final answers~\citep{wang2023selfconsistency,lyu2023faithful}. This highlights a central research challenge: reducing errors by guiding LLMs to identify and follow reliable reasoning paths in long thought chains.

Recent work has sought to optimize the exploration of reasoning branches in LLMs to reduce errors in CoT, by coupling generation with external critics or search procedures, such as Monte Carlo Tree Search (MCTS), to prune faulty trajectories~\citep{li2024hindsight,qin2024o1,xi2024enhancing}. In contrast, Preference Optimization (PO) methods~\citep{lai2024step, zhang2024chain,xu-etal-2025-full} aligned the policy with implicit preferences distilled from reasoning chains, so that decoding itself favors correct trajectories rather than relying on post-hoc filtering.
However, beyond the substantial inference-time overhead of MCTS, these methods share a fundamental limitation that motivates rethinking of the CoT paradigm. Specifically, they all execute CoT reasoning in a strictly step-by-step manner, with each step conditioned on previous ones. Under static teacher-forcing supervision, the model is exposed only to correct prefixes during training, whereas at inference time the prefixes may contain errors that accumulate across steps, resulting in exposure bias that commonly plagues autoregressive generation~\citep{bengio2015scheduled} and imitation learning~\citep{ross2011reduction}.

\begin{figure*}[t]
    \centering
    \includegraphics[width=\textwidth]{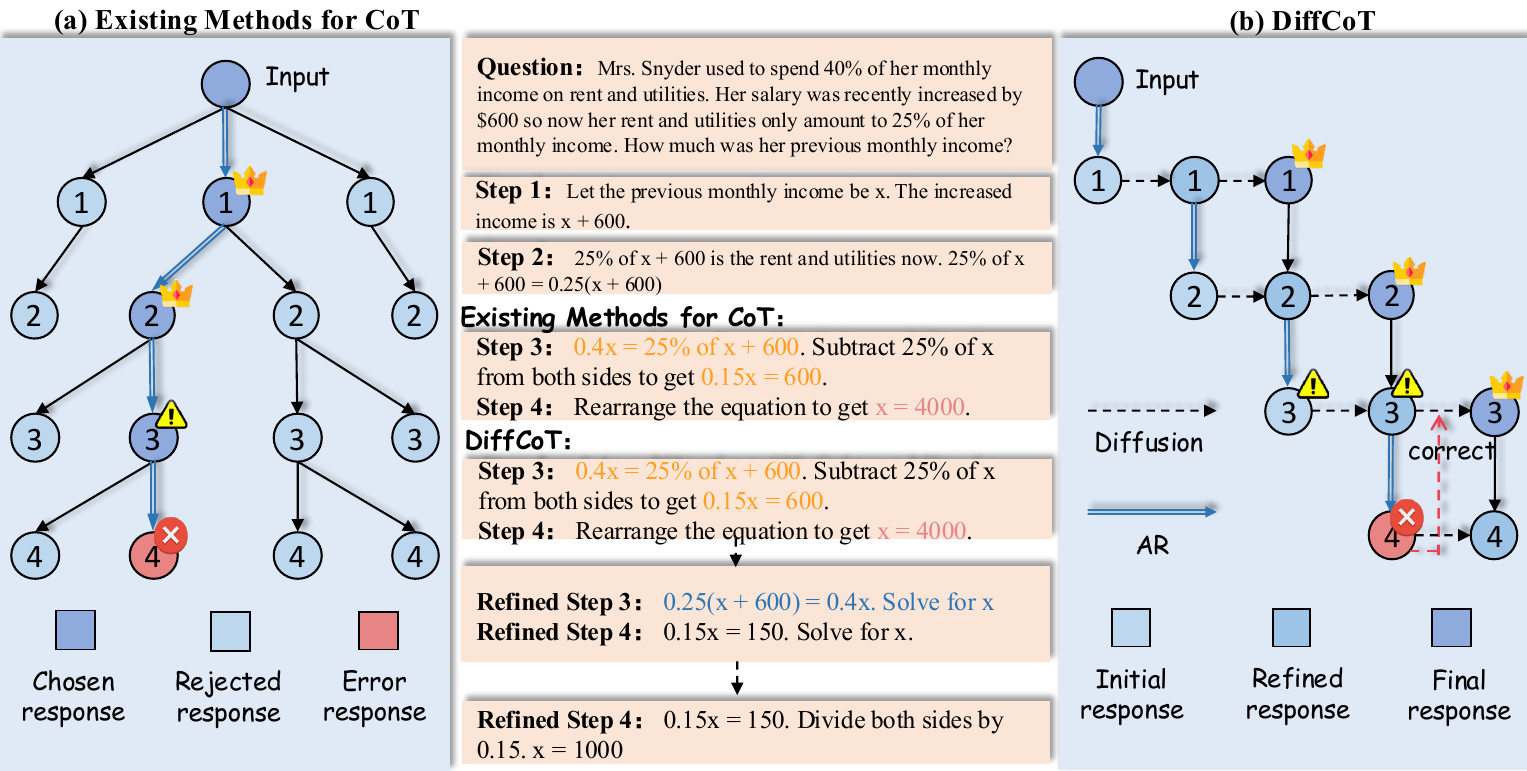}
    \vspace{-0.5cm}
    \caption{
    Comparison of our proposed \textsc{DiffCoT} with existing CoT reasoning approaches:
    \textbf{(a)} Existing step-by-step CoT Reasoning methods adopt teacher-forcing training, where each step depends on the ground-truth output of the previous one. At inference time, this assumption breaks, causing exposure bias and leading to error accumulation.
    \textbf{(b)} \textsc{DiffCoT} performs CoT reasoning along both the noise (diffusion) and temporal (autoregressive) dimensions, enabling iterative correction of prior mistakes and effectively mitigating exposure bias.
    }
    \label{fig:intro}
    \vspace{-0.3cm}
\end{figure*}


To mitigate the limitations above, we formulate CoT reasoning as a globally revisable trajectory, in which intermediate steps are not fixed once generated but can be adaptively corrected in light of later context. {In contrast to autoregressive likelihood maximization, which optimizes conditional factors independently, score-based modeling optimizes a global objective over the full trajectory distribution, implicitly coupling all positions through the gradient field of the log-density.} 
Rather than defining reasoning as a strictly forward, prefix-conditioned process driven by human priors (see Figure~\ref{fig:intro}), our method jointly models all reasoning steps within a diffusion paradigm~\citep{dhariwal2021diffusion}, allowing robust reasoning under realistic noise from a global perspective with long-term revision guided by future signals. Starting from a corrupted or noisy reasoning chain, the entire reasoning trajectory can be iteratively updated, where intermediate revisions and forward progression are seamlessly integrated into a unified refinement process. By unifying iterative revision and forward continuation within the same diffusion-based refinement process, our design philosophy aims to reduce the discrepancy between training-time supervision and inference-time reasoning dynamics, leading to more stable reasoning behavior during inference.

To this end, we propose \underline{Diff}usion-Styled \underline{C}hain \underline{o}f \underline{T}hought, \textbf{\textsc{DiffCoT}}, which integrates diffusion principles into the CoT framework to mitigate the exposure bias of static teacher-forcing and address the pointwise and local-step supervision issues of prior preference optimization solutions. Specifically, 1) \textsc{DiffCoT} modifies only the step-level generation strategy while preserving token-level autoregression, making it straightforward to fine-tune from existing autoregressive models. 2) By incorporating sliding windows and causal noise masks, \textsc{DiffCoT} unifies generation and revision within a versatile framework, to balance autoregressive and diffusion-styled reasoning. 3) Moreover, \textsc{DiffCoT} employs sentence-level noise injection to convert the reasoning paradigm into iterative denoising, progressively refining corrupted trajectories into coherent chains of thought, which captures the distributional evolution of reasoning errors to enable recovery from diverse and compounding mistakes. Altogether, \textsc{DiffCoT} could establish a unified and adaptive paradigm to CoT reasoning in LLMs. 

Our contributions are summarized as follows:
\begin{itemize}[leftmargin=*,nosep]
    \item We propose \textsc{DiffCoT}, a novel diffusion-styled CoT framework that reformulates multi-step reasoning as an iterative denoising process, effectively mitigating exposure bias and error accumulation in autoregressive CoT reasoning.\footnote{The source code is released via \url{https://github.com/caoshidong66/DiffCoT}.}

    \item We introduce a step-level diffusion-based learning strategy that unifies generation and revision through a sliding-window mechanism, enabling retrospective correction of earlier reasoning steps while preserving token-level autoregression.

    \item We design a causal noise schedule that explicitly encodes the temporal dependency of reasoning steps, balancing global error correction with the causal structure required for coherent reasoning.

    \item Extensive experiments on three public mathematical reasoning benchmarks demonstrate that \textsc{DiffCoT} outperforms State-of-The-Art (SoTA) PO methods, yielding robust gains and substantially improved error-correction capability in CoT.
\end{itemize}

\section{Preliminaries}
\paragraph{Chain-of-Thought Reasoning}
Given a question prompt $p$, the CoT paradigm~\citep{wei2022chain} explicitly unfolds the reasoning process into a sequence of reasoning steps $s_{1:K}$ whose final element $s_K$ corresponds to the answer $a$. The conditional distribution of a complete reasoning trace can be expressed as:
{\setlength{\abovedisplayskip}{0.1cm}
\setlength{\belowdisplayskip}{0.1cm}
\begin{equation}
    p_\theta(s_{1:K}\mid p)
    = \prod_{t=1}^{K} \pi_\theta(s_k \mid p, s_{<k}),
\label{equation_1}
\end{equation}}
where $s_k$ denotes the $k$-th reasoning step, $k \le K$ is the step budget, and $\pi_\theta$ is the Auto-Regressive (AR) policy, i.e., the conditional distribution over the next step given the prompt $p$ and the previously generated steps.

On annotated CoT data $(p, s_{1:K})$, the model is trained by maximizing the conditional likelihood:
{\setlength{\abovedisplayskip}{0.1cm}
\setlength{\belowdisplayskip}{0.1cm}
\begin{equation}
\mathcal{L}_{\mathrm{CoT}}
= -\sum_{k=1}^{K}\log \pi_\theta(s_k \mid p, s_{<k}).
\label{equation_2}
\end{equation}}

\paragraph{Diffusion Models}
Diffusion models~\citep{dhariwal2021diffusion} define a generative framework where data samples are gradually corrupted into a tractable noise distribution through a forward process, and a model is trained to approximate the corresponding reverse process. 
Let $x_0 \sim \varphi(x_0)$ denote a data sample from real data distribution $\varphi$. The forward process produces a sequence $\{x_t\}_{t=1}^T$ by applying a noise operator $\mathcal{O}$ at each step:
{\setlength{\abovedisplayskip}{0.1cm}
\setlength{\belowdisplayskip}{0.1cm}
\begin{equation}
x_t = \mathcal{D}(x_{t-1}, \eta_t), \quad t=1,\dots,T,
\label{equation_4}
\end{equation}}
where $t$ denotes the diffusion step index, $\eta_t$ is a noise variable. $\mathcal{O}$ could be a deterministic or structured corruption operator.

The generative process is defined by learning a parameterized reverse transition $p_\theta(x_{t-1} \mid x_t)$ that reconstructs the original sample through stepwise denoising. Training minimizes the discrepancy between the predicted $\hat{x}_{t-1} = f_\theta(x_t, t)$ and the ground-truth corrupted data $x_{t-1}$ obtained from the forward process:
{\setlength{\abovedisplayskip}{0.1cm}
\setlength{\belowdisplayskip}{0.1cm}
\begin{equation}
\begin{split}
&\mathcal{L}_{\mathrm{Diffusion}}
= \\ &\mathbf{E}_{x_0 \sim \varphi(x_0),\, t \sim \{1,\dots,T\}}
\Big[ \ell\big(f_\theta(x_t, t),\, x_{t-1} \big) \Big],
\end{split}
\label{equation_6}
\end{equation}}
where $f_\theta(x_t,t)$ predicts the reconstruction of $x_{t-1}$ from $x_t$, $\ell(\cdot,\cdot)$ is a reconstruction loss. In inference, sampling begins from $x_T$ and iteratively applies the learned reverse transitions until $x_0$ is obtained.

\section{Methodology}

During CoT reasoning, LLMs are typically trained only on correct trajectories~\citep{wei2022chain}, while at inference time they may condition on erroneous intermediate steps~\citep{lyu2023faithful}. Moreover, step-wise optimization focuses on local alignment and ignores future signals within the reasoning trajectory, limiting global consistency~\citep{zhang2024chain}. Together, these issues lead to error accumulation~\citep{yoon2025monte,chen2024diffusion}, motivating a rethinking of CoT reasoning under preference optimization. To address these issues, we argue that effective CoT reasoning requires trajectory-level revision to enforce global consistency and recover from corrupted intermediate steps. As error accumulation leads to noisy reasoning chains at inference time, a mechanism for global recovery becomes essential~\citep{lyu2023faithful,wang2023selfconsistency}. 
Unlike autoregressive decoding that commits to each step irrevocably, diffusion operates through iterative denoising over the entire sequence, enabling global revision and correction of intermediate errors. This property naturally aligns with the need for trajectory-level recovery in CoT reasoning.
Motivated by the robustness of diffusion models in reconstructing structured data from noise~\citep{ho2020denoising,li2022diffusionlm}, we aim to develop a diffusion-styled preference optimization paradigm to reformulate CoT reasoning.



To this end, we propose \textit{\underline{Diff}usion-styled \underline{C}hain \underline{o}f \underline{T}hought} (\textsc{DiffCoT}), which models CoT reasoning in a diffusion-styled framework.
We first present a step-level forward noising process for CoT using reward-ranked candidates (\S\ref{sec_data}), and then apply a diffusion sliding window to iteratively denoise past steps while generating new ones (\S\ref{sec_framework}).
We further introduce causal diffusion noise to strengthen causal consistency across reasoning steps (\S\ref{sec_causal}). An overview is shown in Figure~\ref{fig:method}.

\subsection{Diffusion-Styled Noising for CoT}
\label{sec_data}



Previous preference optimization methods overlooked the distribution between high-reward and low-reward reasoning steps and relied on isolated step-wise adjustments that fail to mitigate the exposure bias inherent in teacher forcing. In this section, we design a diffusion-styled forward noising process for CoT reasoning at the reasoning-step level, where step-level noise is induced by ranking candidate responses for the same step according to their reward scores. In our principle, higher-reward candidates are treated as lower-noise reasoning states, while lower-reward candidates correspond to progressively higher-noise states in the forward process. This ranking forms a progression from clean to corrupted reasoning states under diffusion-styled noise, enabling distribution-aware modeling of reasoning steps.

Specifically, we implement forward noising by first collecting CoT reasoning trajectories for each question to construct the training set $\mathcal{D}$. As shown in Figure~\ref{fig:method}(a), given a question prompt $p$, we follow the standard CoT data generation process by employing MCTS~\citep{browne2012survey}, to gradually build a search tree for step-by-step solution exploration.
The root corresponds to $p$, and each child node represents an intermediate step $s$. A complete path from the root to a terminal node $s_K$ yields a trajectory
$\mathbf{tra}=p\oplus s_1\oplus s_2\oplus \dots \oplus s_K$, where each step $s_k$ is annotated with its reward $r(s_k)$.

In contrast to conventional CoT trajectory construction~\cite{zhang2024chain}, at each step $k$ we collect several candidate responses. These responses are scored using either an external reward model or rollout-based success rates. In our forward noising view, the candidate with the highest reward, denoted as $s^{\sigma^0=w}_{k}$, is regarded as the lowest-noise state for step $k$, and the remaining candidates are ordered by their rewards to form $\{s^{\sigma^{1:T}}_{k}\}$, which we interpret as states with gradually increasing noise.
Step-level noise is thus measured by the deviation of lower-reward responses from this best trajectory, indicating the corruption level at the step granularity.
This design ensures that the final generation remains coherent while providing a set of forward diffusion states for subsequent denoising and preference optimization.

Under the step-level forward noising, \textsc{DiffCoT} defines a diffusion-styled distribution that spans low-noise to high-noise reasoning states~\citep{chen-etal-2025-diffpo}. Note that our data construction does not distinguish between “correct’’ and “incorrect’’ labels at the collection stage as in~\citet{lai2025stepdpo}; instead, it uniformly gathers diverse responses for each step throughout the entire trajectory. The global view could provide distribution-aware data to support both generation and refinement. In this way, error patterns are implicitly encoded in trajectory-level modeling, allowing robust reasoning without additional manual annotations of corrective steps.

\subsection{Reformulating CoT Reasoning as Denoising}
\label{sec_framework}
\begin{figure}[t]
    \centering
    \includegraphics[width=0.85\columnwidth]{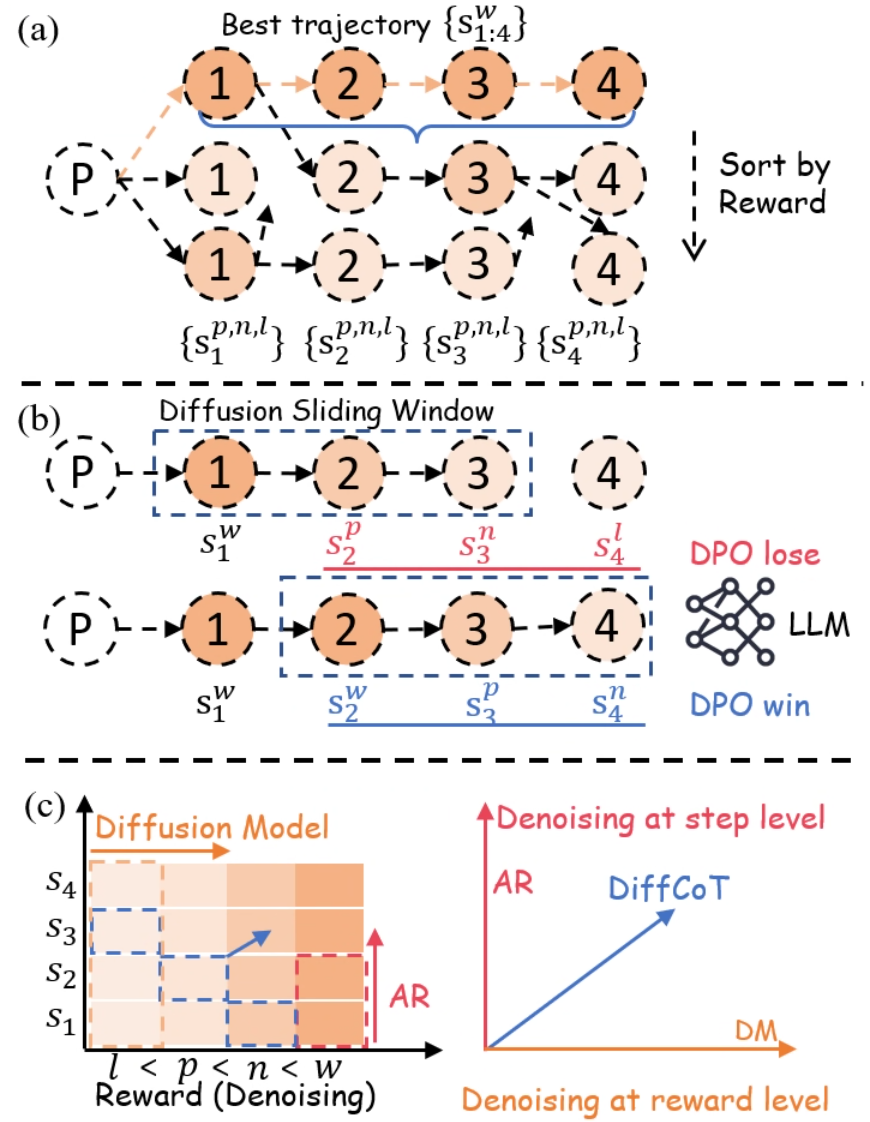}
    \vspace{-0.1cm}
    \caption{
    {\textsc{DiffCoT} Framework and Training Data Construction:}
    \textbf{(a)} Step-level forward noising: MCTS-based data generation defines step-level noise by reward-ranking multiple candidates, yielding states ranging from clean to corrupted.
    \textbf{(b)} Sliding-window denoising: a diffusion sliding window refines previously generated CoT steps while producing the next step in an autoregressive manner.
    \textbf{(c)} Causal diffusion noise: a step-dependent schedule assigns stronger noise to later steps to encode the causal order of the reasoning chain.
    }
    \label{fig:method}
    \vspace{-0.3cm}
\end{figure}

In this section, our \textsc{DiffCoT} framework reformulates CoT reasoning beyond the teacher-forcing paradigm as a diffusion-styled denoising process to mitigate exposure bias at inference time. 
To couple this with the autoregressive nature of CoT, we introduce a diffusion sliding-window mechanism that operates directly on the CoT reasoning process. 
Within the devised sliding window,
previously generated CoT steps are progressively denoised from high-noise toward low-noise reasoning states, thereby facilitating self-correction, while advancing the window naturally aligns with the autoregressive generation of subsequent CoT steps. 
Thus, instead of costly training a separate diffusion language model from scratch, we leverage diffusion-styled data to efficiently fine-tune pre-trained LLM $\pi$ for CoT reasoning.

Specifically, for the iterative denoising process, the model $\pi$ takes the question prompt $p$ together with the preceding reasoning steps $\mathbf{s}_{1:k}$ as input. 
To enable variable-length generation when applying diffusion to thought chains while integrating generation with self-correction, the model maintains a diffusion sliding window of size $m$ and stride $n$. 
At denoising iteration $t$, the window containing previously generated CoT steps $\{s_{k-m}^{\sigma}, \dots, s_k^{\sigma}\}$ is updated to a lower-noise version $\{s_{k-m}^{\sigma'}, \dots, s_k^{\sigma'}\}$, where $\sigma = \sigma_k^{(t)}$ is defined as a function of the step index $k$ and the denoising iteration $t$, while $\sigma$ and $\sigma'$ denote the noise levels before and after refinement, respectively.
Simultaneously, as the window shifts forward by one step, the model predicts the next step $s_{k+1}^{\sigma}$, initialized at a high-noise state. 
Iterating this denoising process eventually yields a clean trajectory $s_{1:K}^{\sigma^0}$:
{\setlength{\abovedisplayskip}{0.1cm}
\setlength{\belowdisplayskip}{0.1cm}
\begin{equation}
\pi_\theta\!\big(p,\, \mathbf{s}_{1:k}^{\,\sigma}\big)
\;\mapsto\;
\underbrace{\mathbf{s}_{k-m:k}^{\,\sigma'}}_{\text{refined past}}
\;\oplus\;
\underbrace{\mathbf{s}_{k+1}^{\,\sigma}}_{\text{predicted future}} \, .
\label{eq:denoise}
\end{equation}}

An illustration of the denoising process is shown in Figure~\ref{fig:method}(b). 
In sequence modeling, teacher-forcing next-token prediction can be viewed as masking along the reasoning-step axis $s$, whereas diffusion corresponds to masking along the noise axis $\sigma$, where $\sigma^{T}$ approaches pure white noise after $T$ iterations. 
To unify both views, we denote $s_k^\sigma$ as the $k$-th CoT step under noise level $\sigma$, where $\sigma$ is instantiated as $\sigma_k^t$, i.e., the diffusion noise strength assigned to the $k$-th step at denoising iteration $t$.

Building on the above reasoning process, we now introduce the training objective in \textsc{DiffCoT}. 
Our goal is to optimize a model $\pi_\theta$ that takes the question prompt $p$ together with the preceding reasoning steps $\mathbf{s}_{1:k}$ as input.

We construct the \textit{win} sequence $s_{k-m:k+1}^{w}$ by combining the denoised steps $\{s_{k-m}^{\sigma'},\dots, s_{k}^{\sigma'}\}$ with a lower-noise variant of $s_{k+1}^{\sigma_{k+1}'}$. In practice, following DiffPO~\citep{chen-etal-2025-diffpo}, we use the best sample $s^{\sigma^{0}}$ as the winning target and consistently align all intermediate generations to this target.
Similarly, the \textit{lose} sequence $s_{k-m:k+1}^{l}$ is formed by combining the unrefined steps $\{s_{k-m}^{\sigma}, \dots, s_{k}^{\sigma}\}$ with the highest-noise variant of $s_{k+1}^{\sigma_{k+1}}$.
The prefix condition is the past text $s_{1:k-1}$. 
To optimize the LLM on this preference pair, we adopt the Direct Preference Optimization (DPO) loss~\citep{rafailov2023direct}:
{\setlength{\abovedisplayskip}{0.1cm}
\setlength{\belowdisplayskip}{0.1cm}
\begin{equation}
\begin{split}
\mathcal{L}_i(\pi_\theta ; \pi_{\textrm{ref}})
= -\log \phi \Big( 
&\beta \log \frac{\pi_\theta(s_{:k+1}^{w} \mid p, s_{1:k})}{\pi_{\textrm{ref}}(s_{:k+1}^{w} \mid p, s_{1:k})} \\
- \beta \log &\frac{\pi_\theta(s_{:k+1}^{l} \mid p, s_{1:k})}{\pi_{\textrm{ref}}(s_{:k+1}^{l} \mid p, s_{1:k})} 
\Big),
\end{split}
\end{equation}}
where $s_{:k+1}$ denotes the subsequence $s_{k-m:k+1}$, $\phi(\cdot)$ denotes the sigmoid function, and $\beta$ is a factor controlling the strength of the preference signal.

Note that our conditional prefix $s_{1:k-1}^{\sigma}$ differs from that used in standard DPO:
instead of being composed exclusively of preferred (i.e., clean) steps,
it combines clean steps from $1$ to $k-m$ with noisy steps within the sliding window. In this manner,
the hybrid construction could alleviate the exposure bias by training the model to make preference-consistent updates
even when conditioned on partially corrupted reasoning prefixes.

\subsection{Causal Diffusion Noise}
\label{sec_causal}

Modeling CoT reasoning with diffusion poses a significant challenge to causality. Conventional full-sequence diffusion models are inherently non-causal~\citep{ho2020denoising,dhariwal2021diffusion}, which contrasts sharply with the causal nature of CoT reasoning. While our backbone $\pi$ retains an AR token-level generation process, prior studies on diffusion models indicate that relying solely on the model’s ability to capture causality is inadequate, especially when reasoning must be performed over noisy or perturbed data~\citep{li2022diffusionlm}.

Inspired by Diffusion Forcing \citep{chen2024diffusion}, as shown in Figure~\ref{fig:method}(c), we leverage noise as a mechanism to inject causal priors into diffusion sliding window.In conventional full-sequence diffusion, the noise strength $\sigma^{(t)}$ depends solely on the denoising iteration $t$ and is shared across all tokens. Such uniform noise injection is ill-suited for CoT reasoning, as it limits the model’s ability to capture step-wise causal dependencies. To overcome this, we redefine the noise schedule as $\sigma^t_k$, a joint function of reasoning step $k$ and iteration $t$ (see \S\ref{sec_framework}). Within the diffusion sliding window, $\sigma^t_k$ follows a progressive schedule, where earlier steps are perturbed with weaker noise, while later steps are perturbed with stronger noise. The noise schedule is formally defined as follows:
{\setlength{\abovedisplayskip}{0.1cm}
\setlength{\belowdisplayskip}{0.1cm}
\begin{equation}
\begin{split}
&F(s_{j-m+1}, \ldots, s_j; j) 
= \\ &\big(s_{j-m+1}^{\sigma^0},\; s_{j-m+2}^{\sigma^{1}},\; \ldots,\; s_j^{\sigma^T}\big),
\end{split}
\end{equation}}
where the diffusion sliding window has size $m$ and stride $1$, and at the $j$-th denoising iteration the window advances to generate $s_j$. In this manner, our framework could better stabilize the causal chain and enhance self-correction in subsequent reasoning steps even if the current step is erroneous.

\section{Experiment}
\begin{table*}[!t]
\centering
\small
\renewcommand{\arraystretch}{1.18}
\setlength{\tabcolsep}{5.2pt}
\resizebox{\linewidth}{!}{
\begin{tabular}{l ccccccc| ccccccc}
\toprule
\textbf{Model} &
\multicolumn{7}{c}{\textbf{Llama3-8B}} &
\multicolumn{7}{c}{\textbf{Qwen3-4B}} \\
\cmidrule(lr){2-8}\cmidrule(lr){9-15}
& CoT & TSFT & CPO & ToT & Step & FStep & \textsc{DiffCoT}
& CoT & TSFT & CPO & ToT & Step & FStep & \textsc{DiffCoT} \\
\midrule
\addlinespace[1.5pt]
\textbf{GSM8K} 
&62.5&62.7&61.6&61.7&63.0&63.2&\textbf{64.4}&84.7&84.5&85.7&85.4&87.5&87.2&\textbf{88.5}\\
\addlinespace[1.5pt]

\textbf{SVAMP}
&72.2&73.8&73.3&73.0&75.7&76.2&\textbf{76.9}&86.8&86.8&87.2&87.9&88.3&89.7&\textbf{90.2} \\
\addlinespace[1.5pt]

\textbf{M-L1}
&43.3&39.1&39.0&39.5&\textbf{43.5}&39.6&39.3&59.3&58.4&60.1&60.9&61.9&64.5&\textbf{75.0}\\
\addlinespace[1.5pt]

\textbf{M-L2}
&29.0&28.4&28.7&29.3&28.9&\textbf{29.7}&26.6&35.0&35.5&35.8&36.0&34.1&39.9&\textbf{48.7} \\
\addlinespace[1.5pt]

\textbf{M-L3}
&13.0&13.6&14.0&13.5&15.7&14.8&\textbf{17.2}&21.6&20.7&21.5&21.8&24.5&27.6&\textbf{33.1}\\
\addlinespace[1.5pt]

\textbf{M-L4}
&7.4&7.8&7.0&7.4&8.0&7.8&\textbf{9.5}&10.5&10.4&11.3&9.5&12.3&18.4&\textbf{24.4} \\
\addlinespace[1.5pt]

\textbf{M-L5}
&2.1&1.7&1.8&1.4&1.9&2.2&\textbf{3.8}&3.9&4.4&4.3&3.2&5.0&5.1&\textbf{13.2}\\
\midrule

\textbf{Model} &
\multicolumn{7}{c}{\textbf{Qwen3-8B}} &
\multicolumn{7}{c}{\textbf{Ministral3-8B}} \\
\cmidrule(lr){2-8}\cmidrule(lr){9-15}
& CoT & TSFT & CPO & ToT & Step & FStep & \textsc{DiffCoT}
& CoT & TSFT & CPO & ToT & Step & FStep & \textsc{DiffCoT} \\
\midrule
\addlinespace[1.5pt]
\textbf{GSM8K}
&87.3&87.5&88.0&86.3&86.4&88.7&\textbf{91.5}& 61.5&61.8&63.8&61.5&66.3&65.9&\textbf{67.4}\\
\addlinespace[1.5pt]

\textbf{SVAMP}
&86.4&87.0&86.4&86.7&87.7&87.9&\textbf{89.3}&82.6&81.6&82.5&82.7&79.1&79.9&\textbf{83.9}\\
\addlinespace[1.5pt]

\textbf{M-L1}
&69.7&69.9&70.4&70.0&71.2&74.3&\textbf{77.0}&50.8&50.6&50.3&50.2&57.2&57.5&\textbf{63.0}\\
\addlinespace[1.5pt]

\textbf{M-L2}
&51.1&51.7&52.0&51.5&51.9&53.2&\textbf{58.1}&30.2&30.1&29.8&30.0&37.0&37.2&\textbf{39.6}\\
\addlinespace[1.5pt]

\textbf{M-L3}
&31.3&30.3&31.6&30.5&32.3&35.4&\textbf{42.8}&22.9&22.5&22.5&22.8&24.7&28.1&\textbf{31.9}\\
\addlinespace[1.5pt]

\textbf{M-L4}
&14.0&14.3&13.5&14.7&14.8&18.8&\textbf{26.6}&6.1&6.8&7.1&7.3&\textbf{14.5}&13.7&11.5\\
\addlinespace[1.5pt]

\textbf{M-L5}
&4.9&4.3&5.0&4.6&5.5&7.1&\textbf{14.9}&2.4&2.9&2.7&2.6&5.3&5.4&\textbf{7.0}\\
\bottomrule
\end{tabular}
}
\vspace{-0.2cm}
\caption{Test reasoning accuracy (\%) on GSM8K, SVAMP, and MATH.
M-L1 to M-L5 denote different difficulty levels in the MATH dataset, where L1 is the easiest and L5 is the most challenging, while Step denotes Step-DPO and FStep denotes Full-Step-DPO.
The best results are in \textbf{bold}.}
\vspace{-0.3cm}
\label{tab:tab1_transposed_2x2}
\end{table*}

\subsection{Settings}
\label{sec:exset}

\paragraph{Models and Datasets} We conduct experiments on four representative backbone models: Llama3-8B \citep{touvron2023llama}, Qwen3-8B \citep{yang2025qwen3}, Ministral3-8B \citep{liu2026ministral} and Qwen3-4B \citep{yang2025qwen3}. We primarily evaluate our \textsc{DiffCoT} on three public mathematical reasoning benchmarks, specifically GSM8K~\citep{cobbe2021training}, SVAMP~\citep{patel2021nlp}, and MATH~\citep{hendrycks2021measuring}. The MATH is divided into five difficulty levels, as defined in the original dataset. More data statistics and preparation are provided in Appendix \S\ref{sec:data_details}.

\vspace{-3pt}
\paragraph{Baselines}
We compare \textsc{DiffCoT} with the following SoTA baselines:  
1)~\textbf{CoT}~\citep{wei2022chain}: generates step-by-step reasoning before the final answer, evaluated with greedy decoding.
2)~\textbf{ToT}~\citep{yao2023tree}: explores multiple reasoning paths via tree search.
3)~\textbf{TS-SFT}~\citep{feng2023alphazero}: applies supervised fine-tuning on reasoning paths obtained from ToT. 
4)~\textbf{CPO}~\citep{zhang2024chain}: performs preference optimization at the step level, directly aligning intermediate reasoning steps via contrastive pairs.
5)~\textbf{Step-DPO}~\citep{lai2025stepdpo}: step-level preference optimization by explicitly collecting erroneous steps and applying preference optimization to correct them.  
\textbf{6) Full-Step-DPO}~\citep{xu-etal-2025-full}: further generalizes step-wise optimization to the entire reasoning trajectory. 
More implementation details regarding model training and data construction are provided in Appendix~\S\ref{sec:imple_details}.

\subsection{Main Results}
\label{exp:main}
As shown in Table~\ref{tab:tab1_transposed_2x2}, \textsc{DiffCoT} improves reasoning performance on a range of modern instruction-tuned language models, such as Qwen3, Ministral3 and Llama3, under standard fine-tuning. This design enables seamless integration into existing training pipelines and allows \textsc{DiffCoT} to deliver consistent performance improvements across models of varying sizes, demonstrating strong scalability across different base models.

From the result, we can observe that: 1) Baselines guided by LLM self-verification signals, such as CPO and ToT, exhibit notable instability across models and datasets, likely because the quality of self-verification is inherently model and task dependent. When the verification signal is poorly calibrated, it may favor plausible but incorrect reasoning paths, leading to inconsistent gains. For example, on Llama3-8B, CPO degrades performance on MATH-L1 but remains competitive on SVAMP.
2) Step-level preference learning baselines, including Step-DPO and Full-Step-DPO, generally achieve great improvements but still suffer from occasional performance drops under certain settings. This suggests that relying solely on local step-wise optimization may not be sufficient to ensure stable reasoning behavior across diverse evaluation scenarios.
3) Overall, \textsc{DiffCoT} outperforms existing PO approaches on most benchmarks and evaluation settings, although it does not achieve the single best result in every case. For example, on MATH-L1 with the Llama3-8B backbone, Step-DPO attains slightly higher accuracy. Beyond this overall trend, we observe that the gains of \textsc{DiffCoT} are more pronounced on stronger backbones, such as Qwen3-8B and Qwen3-4B, suggesting that its more complex reasoning trajectories are better utilized by models with stronger instruction following and reasoning capabilities. We also find that \textsc{DiffCoT} brings larger improvements on harder problems, such as MATH-L4 and MATH-L5, where correct reasoning paths are sparser. This highlights that \textsc{DiffCoT} enables the model to leverage information across different solution paths, making broader exploration more effective, whereas for easier problems, standard CoT-style depth search is often sufficient.

\subsection{Ablation Study}
\label{exp:ablation}

\label{sec:ab}
To validate the effectiveness of the proposed \textsc{DiffCoT} method, we perform an ablation study on its key components. We select two representative base language models from different model families with varied sizes, Llama3-8B and Qwen3-4B, and evaluate them on two general mathematical reasoning benchmarks, GSM8K and SVAMP.

To investigate the impact of incorporating diffusion into CoT reasoning, we evaluate various window sizes to explore how this factor influences the model's performance. Specifically, when the diffusion window size and stride are set to 1, our approach essentially degenerates into an AR method, albeit with differently constructed prefixes. On the other hand, when the sliding diffusion window and stride are set to the number of steps $K$, the method reverts to a purely diffusion-based approach.

As shown in Table~\ref{tab:tab3}, it can be observed that performance degrades when the window size and stride is set to 1 or when it becomes too large. We attribute this phenomenon to a trade-off between causal connectivity and error-correction capability. Strengthening the ability to revise earlier steps often comes at the expense of weakening top-down causal reasoning. When the model operates entirely in AR mode, it exhibits the strongest causal reasoning ability but suffers from exposure bias during testing. In contrast, when the model fully adopts the diffusion mode, an excessively long window introduces noise: it disrupts the structural coherence of reasoning and prolonged denoising fluctuations undermine the causal nature of inference.

\begin{table}[t] \small
    \centering
    \renewcommand{\arraystretch}{1.0}
    \resizebox{0.85\linewidth}{!}{
    \begin{tabular}{l|c|c}
        \toprule
        \textbf{Model} & \textbf{GSM8K (\%)} & \textbf{SVAMP (\%)} \\

        \midrule
        \textbf{Llama3-8B} & \textbf{64.4} & \textbf{76.9} \\
        
        \hspace{1em}- window size, stride=1 
        & 62.9 {\raisebox{0.5ex}{\footnotesize -1.5}} 
        & 75.8 {\raisebox{0.5ex}{\footnotesize -1.1}} \\

        \hspace{1em}- window size, stride=$K$ 
        & 55.4 {\raisebox{0.5ex}{\footnotesize -9.0}} 
        & 68.5 {\raisebox{0.5ex}{\footnotesize -8.4}} \\
        
        \hspace{1em}- causal noise 
        & 62.6 {\raisebox{0.5ex}{\footnotesize -1.8}} 
        & 75.5 {\raisebox{0.5ex}{\footnotesize -1.4}} \\
        
        \midrule
        \textbf{Qwen3-4B} & \textbf{88.5} & \textbf{90.2} \\
        
        \hspace{1em}- window size, stride=1 
        & 87.3 {\raisebox{0.5ex}{\footnotesize -1.2}} 
        & 88.2 {\raisebox{0.5ex}{\footnotesize -2.0}} \\

        \hspace{1em}- window size, stride=$K$
        & 80.0 {\raisebox{0.5ex}{\footnotesize -8.5}} 
        & 82.5 {\raisebox{0.5ex}{\footnotesize -7.7}} \\
        
        \hspace{1em}- causal noise 
        & 86.7 {\raisebox{0.5ex}{\footnotesize -1.8}} 
        & 87.7 {\raisebox{0.5ex}{\footnotesize -2.5}} \\
        
        \bottomrule
    \end{tabular}
    }
    \vspace{-0.2cm}
    \caption{{Ablative results on the general datasets GSM8K and SVAMP, where $K$ is the number of reasoning steps. Numbers shown in the upper-right corner of each cell indicate the relative change in accuracy rate compared to the full \textsc{DiffCoT} model.}}
    \vspace{-0.3cm}
    \label{tab:tab3}
\end{table}

We further conduct an ablation study to verify the effectiveness of the causal diffusion noise. Specifically, we disrupt our proposed noise scheduling by randomly shuffling the data order used for noising, instead of following the accuracy-based progression. This modification effectively breaks the causal structure of the noise. Results in Table~\ref{tab:tab3} show that this ablation leads to substantial performance degradation across different models and datasets, demonstrating that our causal noise scheduling is a critical component of \textsc{DiffCoT}.

\subsection{Analysis}
\label{exp:ana}

\paragraph{Case Study}
To further illustrate the model’s robustness to accumulated imperfections in intermediate reasoning, we present qualitative case studies that contrast effective and suboptimal reasoning trajectories. In particular, we focus on challenging examples in which the model introduces semantically irrelevant or weakly informative steps at an early stage of reasoning. Although such steps are not necessarily incorrect in isolation, they tend to accumulate and hinder progress toward the correct solution under standard step-by-step reasoning.

\begin{figure}[t]
    \centering
    \includegraphics[width=\columnwidth]{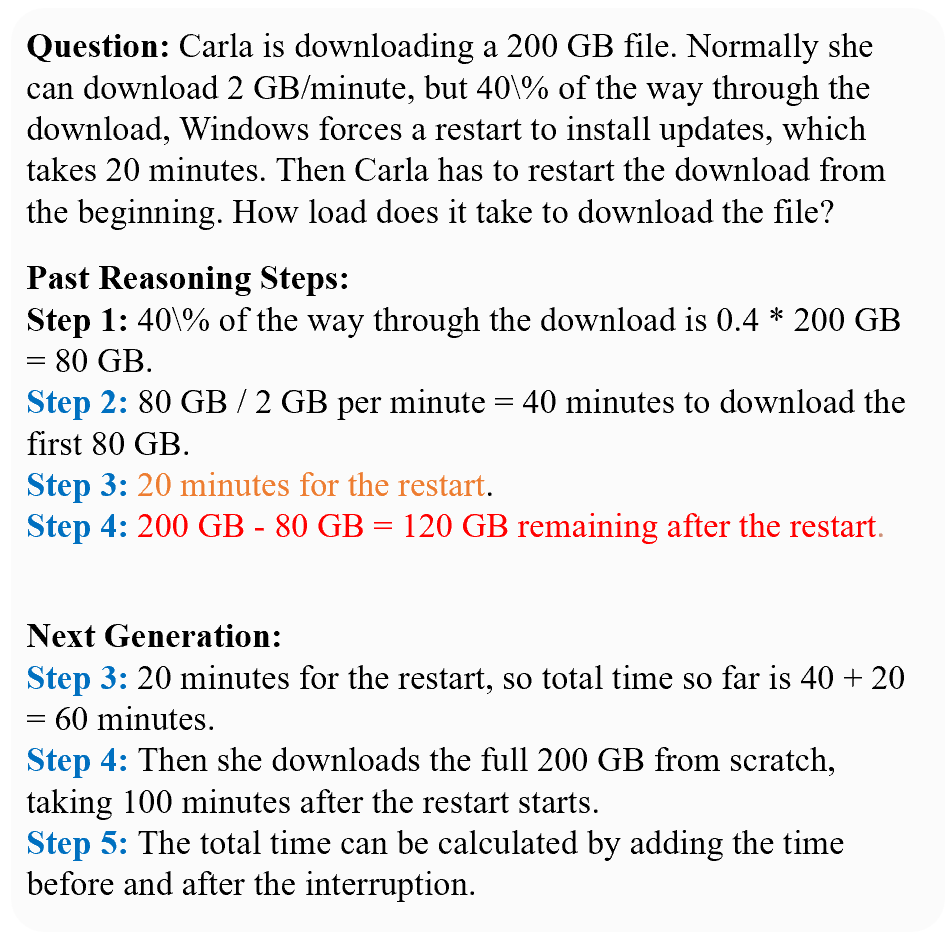}
    \vspace{-0.7cm}
    \caption{Example illustrating how \textsc{DiffCoT} modifies early-stage reasoning shift steps. The steps highlighted in blue represent the diffusion sliding window.}
\label{fig:case_study}
\vspace{-0.3cm}
\end{figure}

As shown in Figure~\ref{fig:case_study}, \textsc{DiffCoT} is able to progressively refine earlier reasoning steps through the diffusion process. Instead of rigidly committing to the initial reasoning prefix, the model gradually improves previously generated steps while producing subsequent ones, effectively revising semantically unhelpful content introduced earlier in the trajectory. This allows the model to move away from locally coherent but globally suboptimal reasoning paths and ultimately reach the correct solution. It highlights a key distinction between \textsc{DiffCoT} and conventional autoregressive reasoning. While autoregressive models tend to propagate early semantic noise forward without revision, \textsc{DiffCoT} leverages its denoising dynamics to refine the entire reasoning trajectory, including previously generated steps, resulting in more robust and globally consistent reasoning behavior.



\begin{figure*}[t]
    \centering
    \includegraphics[width=\textwidth]{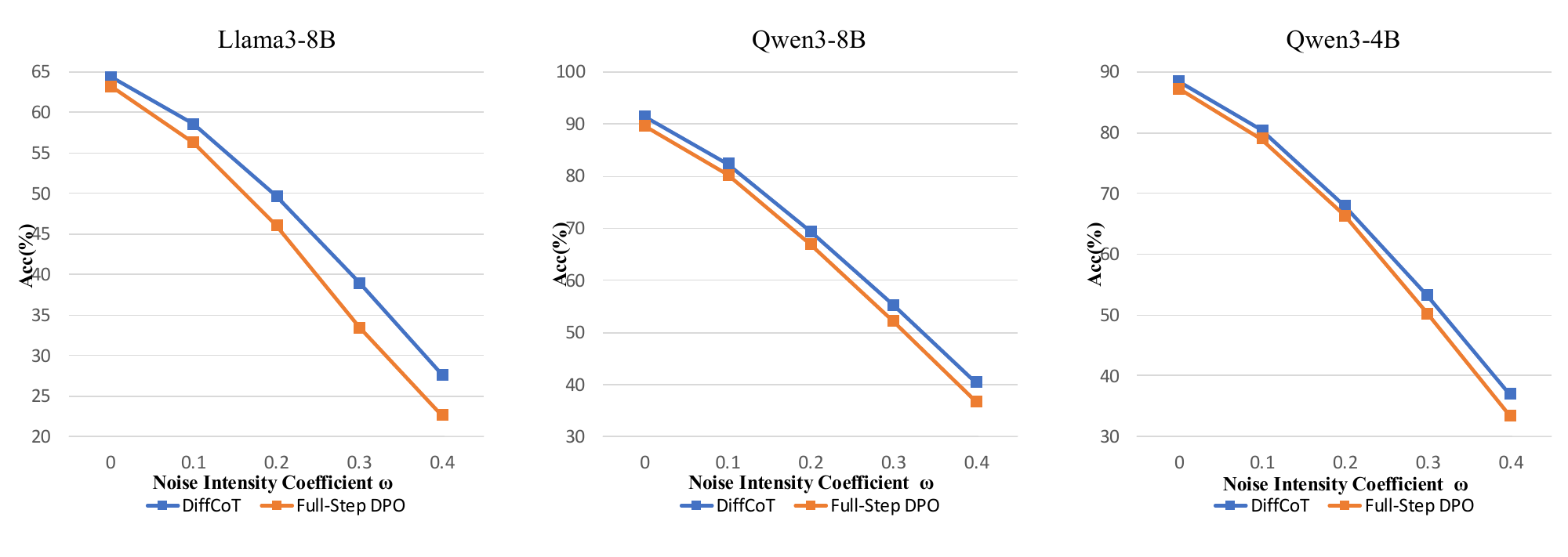}
    \vspace{-0.8cm}
\caption{Correction success rate under stochastic prefix corruption, where noise is injected at the midpoint of the reasoning trajectory with probability $\omega$.
}

    \label{fig:noise}
    \vspace{-0.3cm}
\end{figure*}

\paragraph{Error Accumulation Analysis} 
We further analyze the model’s ability to recover from accumulated imperfections in intermediate reasoning steps. We consider a correction-oriented setting in which the model is deliberately conditioned on prefixes that contain semantically suboptimal or noisy reasoning steps, and is then required to continue the reasoning process to reach a correct final answer. 


To this end, we introduce controlled perturbations at an intermediate stage of the reasoning process. Specifically, when the reasoning reaches approximately half of the trajectory, each preceding step is independently perturbed with probability $\omega$. A perturbation replaces the original step with a low-reward alternative sampled from model-generated trajectories that are semantically plausible but less aligned with the optimal reasoning path. The noise strength $\omega$ therefore governs the degree of accumulated semantic drift in the prefix, enabling a systematic evaluation of robustness under varying levels of intermediate reasoning noise.

We compare different training strategies under this protocol using the correction success rate, defined as the proportion of cases in which the model produces a correct final answer despite being conditioned on a perturbed prefix. Experiments are conducted on 300 GSM8K problems with multiple backbone models, comparing our approach against Full-Step-DPO. As shown in Figure~\ref{fig:noise}, our method consistently achieves substantially higher correction success rates across all settings, demonstrating a stronger ability to compensate for accumulated intermediate noise and to maintain globally coherent reasoning, rather than brittle continuation of imperfect early steps.
In other words, our model can mitigate the impact of early-stage reasoning drift, rather than rigidly propagating locally coherent but globally suboptimal trajectories in CoT reasoning.

\section{Related Work}
\paragraph{Preference Learning}
Preference learning has recently emerged as an effective paradigm for aligning LLMs with human preferences by contrasting desirable and undesirable responses~\citep{ouyang2022training,liu2025lipo}, with methods such as DPO showing strong performance on general language tasks~\citep{rafailov2023direct}. However, extending these gains to mathematical reasoning remains challenging, largely due to the coarse granularity of solution-level supervision, which fails to localize and correct intermediate reasoning errors~\citep{chen-etal-2024-step}. To address this limitation, recent work incorporates step-level preference signals to align intermediate reasoning processes~\citep{lai2024step,lu2024step}, while Full-Step-DPO further optimizes entire reasoning trajectories using global objectives~\citep{xu-etal-2025-full}. Despite these advances, most existing step-wise methods adopt a teacher-forcing paradigm and reason exclusively over clean prefixes during training, leading to error accumulation and degraded robustness during inference. Accordingly, our work reconceptualizes CoT reasoning as a globally revisable trajectory, in which previously generated steps remain malleable to correction in light of future context.



\paragraph{Mathematical Reasoning}
Mathematical reasoning is widely regarded as a challenging capability for large language models, and prior work has explored several directions to improve it. One line of research strengthens base models through continual pre-training on large-scale mathematical corpora or supervised fine-tuning on synthetic datasets distilled from stronger models~\citep{azerbayev2023llemma,deepseek-math,xu-etal-2024-chatglm,mitra2024orcamath,tian2025evolprover}. Another line improves test-time performance by increasing computational budgets, such as generating and reranking multiple solutions using outcome- or process-level rewards, or adopting reward-guided decoding strategies~\citep{pmlr-v267-guan25f,wu2024enhancingmathematicalreasoningllms,wang-etal-2024-math}. Recently, reinforcement learning and preference-based optimization have been explored to directly align reasoning behaviors via trajectory- or step-level supervision, aiming to improve robustness beyond supervised objectives alone~\citep{pal2024smaug,wang-etal-2024-math}. But they largely operate in a forward-only manner and lack mechanisms for revising corrupted intermediate reasoning, motivating our diffusion-styled reformulation of CoT reasoning to enable effective error correction.


\section{Conclusion and Future Work}
In this work, we proposed a novel \textsc{DiffCoT} framework to enhance mathematical reasoning and alleviate error accumulation by integrating diffusion steps into autoregressive generation. Our method utilizes a sliding diffusion window with causal noise to refine intermediate steps while maintaining consistent reasoning chains. Extensive experiments confirm that our proposed \textsc{DiffCoT} outperforms existing CoT methods, offering a robust solution for multi-step CoT reasoning. Future work will explore the scalability of our paradigm across more backbone models and broader reasoning domains.
\section*{Limitations}
Although we have conducted extensive experiments on \textsc{DiffCoT} and observed strong empirical performance, several limitations remain. First, the data construction and training of \textsc{DiffCoT} follow an off-policy paradigm, where preference data are collected using a policy that differs from the one being optimized. Such a mismatch between the behavior policy and the training policy may introduce distribution shift, biased value estimation, and training instability, especially when scaling to more difficult datasets or longer reasoning chains. Second, similar to Diffusion Forcing, \textsc{DiffCoT} breaks the local Markov property of prefix conditioned generation by revisiting and modifying historical reasoning steps. While this violation enables stronger capabilities, it also increases the uncertainty and controllability challenges during generation, typically requiring more training iterations and larger amounts of data to achieve stable convergence.


\section*{Acknowledgments}

This work is partially supported by National Natural Science Foundation of China Young Scientists Fund (No. 62206233) and RMGS (2025 First Processing Cycle).

\bibliography{custom}
\newpage
\appendix

\section{Data Statistics}
\label{sec:data_details}
We primarily evaluate our \textsc{DiffCoT} on mathematical reasoning benchmarks, specifically SVAMP~\citep{patel2021nlp}, MATH~\citep{hendrycks2021measuring} and GSM8K~\citep{cobbe2021training}. The MATH is divided into five difficulty levels, as defined in the original dataset. To reduce randomness and better showcase the experimental results, we sample test instances separately from each of five levels.
To maintain a reasonable computational budget, particularly given the high cost of Tree-of-Thought~\citep{yao2023tree} search, we restrict each dataset to at most 300 test samples through random sampling. 
For training, we similarly sample 300 instances from each dataset to construct the training set. For a fair comparison, all baseline methods use the same dataset configuration. We provide the detailed data statistics in Table~\ref{tab:statistics}.

\begin{table}[h]
\centering
\small
\resizebox{\linewidth}{!}{
\begin{tabular}{l|cccc}
\toprule
Dataset  & \#Train & \#Test &\#Step $K$ &\#window \\
\midrule
GSM8K   & 7473  & 1319 & 5 & 2\\
SVAMP   & 794 & 206 & 3 & 2\\
MATH-L1 & 564 & 437 & 3 & 2\\
MATH-L2 & 1348 & 894  & 4 & 2\\
MATH-L3 & 1592 & 1131  & 5 & 2\\
MATH-L4 & 1690 & 1214  & 6 & 3\\
MATH-L5 & 2304 & 1324  & 6 & 3\\
\bottomrule
\end{tabular}}
\caption{Dataset-specific configurations for the reasoning experiments. 
$\#\mathrm{Train}$ and $\#\mathrm{Test}$ denote the numbers of training and test samples, respectively. 
$\#\mathrm{Step}\ K$ denotes the maximum number of reasoning steps, while $\#\mathrm{window}$ denotes the number of reasoning windows.}
\label{tab:statistics}
\end{table}

\section{Implementation Details}
\label{sec:imple_details}

\paragraph{Data Generation via Rollout-Based Thought Search}

We generate candidate contexts by conducting an iterative thought search starting from each original problem. 
Specifically, at reasoning step $t$, we maintain a prefix that contains all previously selected thoughts.

Conditioned on the current prefix, we sample five candidate thoughts from the model. 
Figure~\ref{fig:dataset_example} provides a representative example illustrating the constructed candidate thoughts and their corresponding step-wise evaluations under this procedure.

To evaluate each candidate thought, we estimate its utility via Monte-Carlo rollouts. 
Concretely, we perform $R=8$ independent rollouts conditioned on the concatenation of the prefix and the candidate thought. 
Each rollout is decoded until termination to obtain a final answer. 
We then compute the empirical success rate as the fraction of rollouts that yield the correct answer.

We continue the search by greedily selecting the candidate with the highest success rate, appending it to the prefix, 
and repeating the above procedure until the reasoning process terminates. We provide a detailed example with step-wise reasoning annotations in Figure~\ref{fig:dataset_example}.

\paragraph{Model Training}
For efficient fine-tuning, we employ Low-Rank Adaptation (LoRA)~\citep{hu2022lora} with a rank of 8 and $\alpha=16$, where LoRA adapters are inserted into the \texttt{q\_proj} and \texttt{v\_proj} linear projections of every self-attention layer. Model training is conducted using the DPO~\citep{rafailov2023direct} loss with a regularization coefficient of $\beta=0.1$, optimized by AdamW~\citep{loshchilov2017decoupled} with a cosine learning rate schedule.We train for only one epoch. For the MATH dataset, we use a learning rate of $1\times10^{-5}$ for Qwen3-8B, $2\times10^{-5}$ for Qwen3-4B, $1.5\times10^{-5}$ for Llama3-8B, and $2.5\times10^{-5}$ for Ministral. For SVAMP and GSM8K, we use a learning rate of $2\times10^{-5}$. We set the warm-up ratio to 0.05 and the global batch size to 8. During decoding, the temperature is fixed at 0.
Compared results (p < 0.05 under t-test) are averaged over three random runs. All experiments are conducted on NVIDIA A100 GPUs 80GB. Training and inference on 300 GSM8K samples require approximately 11 GPU-hours in total.

\paragraph{Baseline Implementation} The instruction-tuned backbone models are implemented by adopting the `Meta-Llama-3-8B-Instruct\footnote{\url{https://huggingface.co/meta-llama/Meta-Llama-3-8B-Instruct}}', `Qwen3-8B\footnote{\url{https://huggingface.co/Qwen/Qwen3-8B}}', `Ministral3-8B-Instruct-2512\footnote{\url{https://huggingface.co/mistralai/Ministral-3-8B-Instruct-2512-BF16}}' and `Qwen3-4B-Instruct-2507\footnote{\url{https://huggingface.co/Qwen/Qwen3-4B-Instruct-2507}}' versions. 

For Full-Step-DPO, we do not retrain the Process Reward Model (PRM) introduced in Full-Step-DPO. In our implementation, we retain the proposed global step-wise loss formulation, but replace the PRM-based step rewards with empirical success rates estimated from Monte-Carlo rollouts. Specifically, we collect extra complete reasoning trajectories from rollouts and use their final outcomes to compute success-based step scores, which serve as substitutes for PRM outputs in the global loss. 

Moreover, for CPO and ToT baselines, we follow the implementation described in the CPO paper, where preference pairs are constructed via LLM self-evaluation of intermediate reasoning steps. This design choice highlights the fundamental distinction between CPO-style self-judgment-based supervision and Step-DPO-style supervision derived from execution or outcome feedback.

\section{Discussion with Reinforcement Learning}
PO methods generally follow a paradigm: trajectory generation followed by trajectory-level reweighting. The model first expands a full reasoning path via longitudinal generation. Alignment is then applied horizontally across completed trajectories using either reinforcement learning or preference optimization, discouraging low-reward reasoning paths and favoring high-reward ones.

While this paradigm of vertical generation followed by horizontal RL-based refinement has achieved broad empirical success, it inevitably suffers from error accumulation~\cite{gan2025rethinking, tian2025codehalu, sun2026factecausalityinspiredevaluationtrustworthy}, as early mistakes in the generated trajectory can propagate and constrain subsequent optimization.

\citet{chen-etal-2025-diffpo} theoretically showed that the denoising process of diffusion models, which gradually transforms low-quality samples into high-quality ones, is equivalent to preference optimization. Building on this insight, \textsc{DiffCoT} introduces a diffusion sliding window, tightly coupling longitudinal reasoning generation with horizontal preference optimization within a unified framework.

From this perspective, we believe that future extensions of our method can naturally integrate reinforcement learning, leveraging RL to directly optimize the denoising dynamics of the diffusion sliding window.

\section{Additional Experiments}
In this section, we present additional experiments to further evaluate our method. Specifically, we analyze the sensitivity of parameter $\beta$ in~\S\ref{sec:beta}, report results on the full test sets in~\S\ref{sec:fulltest}, examine performance under different evaluation protocols in~\S\ref{sec:eval_protocol}, and provide additional results on AIME in~\S\ref{sec:aime}.

\subsection{Sensitivity Analysis of Parameter $\beta$}
\label{sec:beta}
We analyze the sensitivity of the $\beta$ parameter in the DPO loss, which controls the difference between the new and old policies. Specifically, we conduct this experiment with Qwen3-4B and Llama3-8B on the SVAMP and MATH-L4 benchmarks.

As shown in Table~\ref{tab:tab5}, the results show that our method remains stable within a reasonable variation range, without suffering a sharp performance drop, which demonstrates its robustness.

\begin{table}[!htbp] \small
    \centering
    \renewcommand{\arraystretch}{1.0}
    \resizebox{0.85\linewidth}{!}{
    \begin{tabular}{l|c|c}
        \toprule
        \textbf{Model} & \textbf{SVAMP (\%)} & \textbf{MATH-L4 (\%)} \\

        \midrule
        \textbf{Llama3-8B} &  & \\
        
        \hspace{1em}- $\beta=0.05$ 
        & 75.0 {\raisebox{0.5ex}{\footnotesize -1.9}} 
        & 8.3 {\raisebox{0.5ex}{\footnotesize -1.2}} \\

        \hspace{1em}- $\beta = 0.1$
        & \textbf{76.9}
        & \textbf{9.5}  \\
        
        \hspace{1em}- $\beta = 0.15$
        & 76.5 {\raisebox{0.5ex}{\footnotesize -0.4}} 
        & 9.0 {\raisebox{0.5ex}{\footnotesize -0.5}} \\
        \hspace{1em}- $\beta = 0.2$
        & 74.7 {\raisebox{0.5ex}{\footnotesize -1.8}} 
        & 8.7 {\raisebox{0.5ex}{\footnotesize -0.8}} \\

        \midrule
        \textbf{Qwen3-4B} & & \\
        
        \hspace{1em}- $\beta=0.05$ 
        & 87.2 {\raisebox{0.5ex}{\footnotesize -3.0}} 
        & 22.1 {\raisebox{0.5ex}{\footnotesize -2.3}} \\

        \hspace{1em}- $\beta=0.1$ 
        & \textbf{90.2} 
        & \textbf{24.4} \\
        
        \hspace{1em}- $\beta=0.15$ 
        & 89.9 {\raisebox{0.5ex}{\footnotesize -0.3}} 
        & 24.2 {\raisebox{0.5ex}{\footnotesize -0.2}} \\
        
        \hspace{1em}- $\beta=0.2$ 
        & 88.8 {\raisebox{0.5ex}{\footnotesize -1.4}} 
        & 22.9 {\raisebox{0.5ex}{\footnotesize -1.5}} \\
        
        \bottomrule
    \end{tabular}
    }
    \vspace{-0.2cm}
    \caption{Sensitivity Analysis of Parameter $\beta$. The best results are highlighted in \textbf{bold}. Numbers shown in the upper-right corner of each cell indicate the relative change in accuracy rate compared to the default $\beta = 0.1$ setting.}
    \label{tab:tab5}
\end{table}

\subsection{Results on the Full Test Set}
\label{sec:fulltest}
To mitigate the potential impact of dataset splitting, we additionally evaluate our method on the full test sets. Specifically, we use Llama3-8B, Qwen3-4B, and Qwen3-8B for evaluation on the full SVAMP test set and the full MATH-L4 test set.

\begin{table}[!htbp] \small
    \centering
    \renewcommand{\arraystretch}{1.0}
    \resizebox{0.85\linewidth}{!}{
    \begin{tabular}{l|c|c}
        \toprule
        \textbf{Method} & \textbf{SVAMP (\%)} & \textbf{MATH-L4 (\%)} \\

        \midrule
        \multicolumn{3}{c}{Llama3-8B} \\
        \midrule
        CoT & 74.7 & 6.0\\
        
        Step-DPO
        & 75.8 
        & \textbf{8.6}  \\

       Full-Step-DPO 
        & 75.9 
        & 7.4  \\
        
        \textsc{DiffCoT}
        & \textbf{77.1} 
        & 7.8  \\
        
        \midrule
        \multicolumn{3}{c}{Qwen3-4B} \\
        \midrule
        CoT & 80.5 & 8.2 \\
        
        Step-DPO
        & 88.5 & 9.4 \\

        Full-Step-DPO
        & 87.3 & 15.6\\
        
        \textsc{DiffCoT}
        & \textbf{90.2} & \textbf{22.6}\\

        \midrule
        \multicolumn{3}{c}{Qwen3-8B} \\
        \midrule
        CoT &85.7 & 12.6\\
        
        Step-DPO
        & 85.9 & 12.8 \\

        Full-Step-DPO
        & 87.1 & 17.0 \\
        
        \textsc{DiffCoT}
        & \textbf{89.3} &\textbf{23.2} \\

        \bottomrule
    \end{tabular}
    }
    \vspace{-0.2cm}
    \caption{Results on the full SVAMP and MATH-L4 test sets. To mitigate the potential impact of dataset splitting, we additionally evaluate our method using Llama3-8B, Qwen3-4B, and Qwen3-8B on the full test sets. The best results are highlighted in \textbf{bold}.}
    \vspace{-0.3cm}
    \label{tab:tab6}
\end{table}

As shown in Table~\ref{tab:tab6}, our model achieves the best performance on most model--dataset combinations. In particular, on the Qwen models with MATH-L4, it substantially outperforms previous methods.

\subsection{Results under Different Evaluation Protocols}
\label{sec:eval_protocol}
To further assess the generation quality of \textsc{DiffCoT}, we evaluate it under multiple evaluation protocols. In the main paper, we adopt a strict criterion, where a prediction is counted as correct only if the model outputs the correct numerical answer in the required format. To better reflect the actual reasoning ability of \textsc{DiffCoT}, we further consider two additional protocols: \textit{format\_agnostic}, which ignores the formatting requirement and extracts the last occurring number in the output as the predicted answer, and \textit{contains\_gt}, which checks whether the generated text contains the ground-truth answer anywhere in the response. 

\begin{table}[!htbp] \small
    \centering
    \renewcommand{\arraystretch}{1.0}
    \resizebox{0.85\linewidth}{!}{
    \begin{tabular}{l|c|c|c}
        \toprule
        \textbf{Method} & \textbf{Strict (\%)} & \textbf{ Agnostic(\%)} & \textbf{Contain (\%)} \\

        \midrule
        \multicolumn{4}{c}{GSM8K} \\
        \midrule
        CoT & 84.7& 86.1 & 88.7\\
        
        Step-DPO
        & 87.5& 88.9
        & 90.7  \\

       FStep
        & 87.2 & 88.4
        & 89.1 \\
        
        \textsc{DiffCoT}
        & \textbf{88.5} & \textbf{90.8}
        & \textbf{91.6}  \\
        
        \midrule
        \multicolumn{4}{c}{SVAMP} \\
        \midrule
        CoT &86.8 &87.8 & 89.2 \\
        
        Step-DPO
        &88.3 & 89.3& \textbf{94.7} \\

        FStep
        & 89.7&90.2 & 92.0\\
        
        \textsc{DiffCoT}
        & \textbf{90.2}&\textbf{90.8} & 92.2\\

        \midrule
        \multicolumn{4}{c}{MATH-L1} \\
        \midrule
        CoT &59.3&62.7 & 88.3\\
        
        Step-DPO
        & 61.9& 66.3& 89.0 \\

        FStep
        &64.5 &67.1 & 88.7 \\
        
        \textsc{DiffCoT}
        & \textbf{75.0}& \textbf{88.0}&\textbf{91.7} \\

        \midrule
        \multicolumn{4}{c}{MATH-L2} \\
        \midrule
        CoT &35.0&38.9 & 75.6\\
        
        Step-DPO
        &34.1 &38.8 & 75.4 \\

        FStep 
        & 39.9&41.7 & 76.0 \\
        
        \textsc{DiffCoT}
        & \textbf{48.7}&\textbf{77.0} &\textbf{87.7} \\

        \midrule
        \multicolumn{4}{c}{MATH-L3} \\
        \midrule
        CoT &21.6&26.2& 71.7\\
        
        Step-DPO
        & 24.5&25.6 & 75.4 \\

        FStep
        & 27.6&29.1 & 77.0 \\
        
        \textsc{DiffCoT}
        & \textbf{33.1}&\textbf{73.8} &\textbf{86.3} \\

        \midrule
        \multicolumn{4}{c}{MATH-L4} \\
        \midrule
        CoT &10.5&11.6 & 59.6\\
        
        Step-DPO
        & 12.3&13.0 & 59.4 \\

        FStep
        & 18.4&17.6 & 62.1 \\
        
        \textsc{DiffCoT}
        & \textbf{24.4}&\textbf{62.9} &\textbf{78.3} \\

        \midrule
        \multicolumn{4}{c}{MATH-L5} \\
        \midrule
        CoT &3.9&6.8 & 48.0\\
        
        Step-DPO
        & 5.0&7.3 & 51.7 \\

        FStep
        & 5.1&8.4 & 58.9 \\
        
        \textsc{DiffCoT}
        & \textbf{13.2}&\textbf{37.3} &\textbf{61.9} \\

        \bottomrule
    \end{tabular}
    }
    \vspace{-0.2cm}
\caption{Qwen3-4B results under different evaluation protocols. The best results are highlighted in \textbf{bold}. FStep denotes Full-Step-DPO. Agnostic  denotes \textit{format\_agnostic}, and Contain  denotes \textit{contains\_gt}.}
    \vspace{-0.3cm}
    \label{tab:tab7}
\end{table}

The results of Qwen3-4B and Qwen3-8B on different datasets under these evaluation protocols are shown in Table~\ref{tab:tab7} and Table~\ref{tab:tab8}. As can be seen, \textsc{DiffCoT} consistently achieves clear improvements under all three evaluation protocols. This indicates that the gains of our method do not merely come from better instruction following or answer formatting, but reflect a real enhancement in the model's underlying reasoning capability.

\begin{table}[!htbp] \small
    \centering
    \renewcommand{\arraystretch}{1.0}
    \resizebox{0.85\linewidth}{!}{
    \begin{tabular}{l|c|c|c}
        \toprule
        \textbf{Method} & \textbf{Strict (\%)} & \textbf{ Agnostic(\%)} & \textbf{Contain (\%)} \\

        \midrule
        \multicolumn{4}{c}{GSM8K} \\
        \midrule
        CoT & 87.3& 88.1 &90.8 \\
        
        Step-DPO
        & 86.4& 87.6
        & \textbf{93.4}  \\

       FStep
        & 88.7 & 89.0
        & 93.3 \\
        
        \textsc{DiffCoT}
        & \textbf{91.3} & \textbf{91.4}
        & 92.9  \\
        
        \midrule
        \multicolumn{4}{c}{SVAMP} \\
        \midrule
        CoT &86.4 &88.3 & 91.7 \\
        
        Step-DPO
        &87.7 & 88.9& 92.1 \\

        FStep 
        &87.9&89.4 & 92.3\\
        
        \textsc{DiffCoT}
        & \textbf{89.3}&\textbf{89.8} & \textbf{93.6}\\

        \midrule
        \multicolumn{4}{c}{MATH-L1} \\
        \midrule
        CoT &69.7&75.6 & 87.0\\
        
        Step-DPO
        & 71.2& 76.7&88.0\\

        FStep 
        &74.3&78.5 & \textbf{88.3} \\
        
        \textsc{DiffCoT}
        & \textbf{77.0}& \textbf{79.9}&85.1 \\

        \midrule
        \multicolumn{4}{c}{MATH-L2} \\
        \midrule
        CoT &51.1&58.0 & 81.3\\
        
        Step-DPO
        &51.9 &59.4 & \textbf{82.2} \\

        FStep 
        & 53.2&61.7 & 81.9 \\
        
        \textsc{DiffCoT}
        & \textbf{58.1}&\textbf{74.3} &78.4 \\

        \midrule
        \multicolumn{4}{c}{MATH-L3} \\
        \midrule
        CoT &31.3&37.0& 77.8\\
        
        Step-DPO
        & 32.3&37.2& 78.1 \\

        FStep 
        & 35.4&38.0 & 78.3 \\
        
        \textsc{DiffCoT}
        & \textbf{42.8}&\textbf{71.5} &\textbf{79.0} \\

        \midrule
        \multicolumn{4}{c}{MATH-L4} \\
        \midrule
        CoT &14.0&16.4 & 62.4\\
        
        Step-DPO
        & 14.8&16.5 & 62.7 \\

        FStep 
        & 18.8&19.4 & 63.9 \\
        
        \textsc{DiffCoT}
        & \textbf{26.6}&\textbf{43.9} &\textbf{70.1} \\

        \midrule
        \multicolumn{4}{c}{MATH-L5} \\
        \midrule
        CoT &4.9&6.9 & 51.2\\
        
        Step-DPO
        & 5.5&7.4 & 52.8 \\

        FStep 
        & 7.1&8.6 & 59.2 \\
        
        \textsc{DiffCoT}
        & \textbf{14.9}&\textbf{49.0} &\textbf{65.7} \\

        \bottomrule
    \end{tabular}
    }
    \vspace{-0.2cm}
\caption{Qwen3-8B results under different evaluation protocols. The best results are highlighted in \textbf{bold}. FStep denotes Full-Step-DPO. Agnostic  denotes \textit{format\_agnostic}, and Contain  denotes \textit{contains\_gt}.}
    \vspace{-0.3cm}
    \label{tab:tab8}
\end{table}

\subsection{Results on AIME}
\label{sec:aime}
To further evaluate the effectiveness of our method on more challenging mathematical reasoning tasks, we additionally conduct experiments on the AIME benchmark~\citep{maa_aime}. Specifically, we train our method on AIME problems from 1983 to 2024 and evaluate it on AIME 2025. We use the experimental setting of step = 6 and window = 3, while keeping all other hyperparameters the same as in the other experiments.

However, since the base reasoning abilities of Llama3-8B and Ministral3-8B are too weak, both models achieve zero correct answers on AIME 1983--2025. As a result, even with multiple rollouts, they produce only a very limited number of correct response samples, which makes preference optimization training infeasible. Therefore, we only report the results of Qwen3-4B and Qwen3-8B.

As shown in Table~\ref{tab:tab9}, our method remains competitive on this benchmark and achieves strong performance across different backbones. These results further demonstrate that our method generalizes well to harder mathematical reasoning benchmarks beyond SVAMP and MATH.

\begin{table}[!htbp] \small
    \centering
    \renewcommand{\arraystretch}{1.0}
    \resizebox{0.85\linewidth}{!}{
    \begin{tabular}{l|c|c|c}
        \toprule
        \textbf{Method} & \textbf{Strict (\%)} & \textbf{ Agnostic(\%)} & \textbf{Contain (\%)} \\

        \midrule
        \multicolumn{4}{c}{Qwen3-4B} \\
        \midrule
        CoT & 5.5& 6.6 &23.3 \\
        
        Step-DPO
        & 10.0& 10.0
        & 28.8  \\

       FStep
        & 12.2 & 13.3
        & 30.0 \\
        
        \textsc{DiffCoT}
        & \textbf{30.0} & \textbf{32.2}
        & \textbf{35.5}  \\
        
        \midrule
        \multicolumn{4}{c}{Qwen3-8B} \\
        \midrule
        CoT & 8.8 &10.0 &17.7  \\
        
        Step-DPO
        &  14.4 & 14.4 & 24.4 \\

        FStep 
        & 12.2 &13.3 & 20.0 \\
        
        \textsc{DiffCoT}
        &\textbf{20.0} &\textbf{20.0} &\textbf{26.6} \\

        \bottomrule
    \end{tabular}
    }
    \vspace{-0.2cm}
\caption{Result on AIME Benchmark. The best results are highlighted in \textbf{bold}, FStep denotes Full-Step-DPO. Agnostic  denotes \textit{format\_agnostic}, and Contain  denotes \textit{contains\_gt}.}
    \vspace{-0.3cm}
    \label{tab:tab9}
\end{table}

\section{Generative AI Usage}
AI assistants were used in a limited and supportive manner during the preparation of this manuscript, primarily for language polishing, formatting suggestions, and improving clarity of presentation. 

\begin{figure*}[!t]
\centering
\begin{tcolorbox}[
  width=\textwidth,
  colback=gray!3,
  colframe=black!40,
  boxrule=0.3pt,
  arc=2pt
]

\footnotesize

\textbf{Question.}~
In a truck, there are 26 pink hard hats, 15 green hard hats, and 24 yellow hard hats.
If Carl takes away 4 pink hard hats, and John takes away 6 pink hard hats and twice as many green hard hats
as the number of pink hard hats that he removed, calculate the total number of hard hats that remained in the truck.

\medskip
\textbf{Prompt.}~
You are an expert math problem solver. You must reason step by step and avoid logical or arithmetic mistakes.
Solve the following math problem. You MUST use exactly 5 reasoning steps. After the reasoning, output the final answer
in the last line using the format: \texttt{Answer: <final\_answer>}.

\medskip
\textbf{Step 1.}
\begin{itemize}\setlength\itemsep{2pt}
  \item[{\small[0]}] 26 pink hard hats $-$ 4 pink hard hats $=$ 22 pink hard hats \hfill \texttt{success\_rate=0.0}
  \item[{\small[1]}] 26 $-$ 4 $=$ ? (number of pink hats after Carl's removal) \hfill \texttt{success\_rate=0.125}
  \item[{\small[2]}] Initial total number of hard hats: 26 + 15 + 24 = 65.   \hfill \texttt{success\_rate=0.5}
  \item[{\small[3]}] Initial number of pink hard hats is 26. 
  \hfill \texttt{success\_rate=0.5}
  \item[{\small[4]}] We have 26 pink hard hats initially. Carl takes away 4 pink hard hats, leaving us with $26-4=22$ pink hard hats.
  \hfill \texttt{success\_rate=0.75}

\end{itemize}

\medskip
\textbf{Step 2.}
\begin{itemize}\setlength\itemsep{2pt}
  \item[{\small[0]}] John removed 6 pink hats and 6 green hats, leaving $15-6=9$ green hats. \hfill \texttt{success\_rate=0.0}
  \item[{\small[1]}] John removes 6 pink hats, so he removes 6 green hats as well. Thus $15-6=9$ green remain.
  \hfill \texttt{success\_rate=0.125}
  \item[{\small[2]}] Carl removes 4 pink hard hats: $26 - 4 = 22$.
 \hfill \texttt{success\_rate=0.5}
  \item[{\small[3]}] 6 pink hats are removed by John, leaving $22-6=16$ pink hats. \hfill \texttt{success\_rate=0.625}
  \item[{\small[4]}] John removes 6 pink hats: $22-6=16$ pink left. He removes twice as many green hats: $2\times 6=12$,so $15-12=3$ green left.
  \hfill \texttt{success\_rate=1.0}

\end{itemize}

\medskip
\textbf{Step 3.}
\begin{itemize}\setlength\itemsep{2pt}
  \item[{\small[0]}] 24 yellow hard hats remain in the truck \hfill \texttt{success\_rate=0.75}
  \item[{\small[1]}] We still have 24 yellow hard hats. \hfill \texttt{success\_rate=0.75}
  \item[{\small[2]}] Yellow hard hats remain unchanged, so we still have 24 yellow hard hats. \hfill \texttt{success\_rate=0.875}
  \item[{\small[3]}] 24 yellow hard hats remain in the truck and are not affected by the removals. \hfill \texttt{success\_rate=0.875}
 \item[{\small[4]}] 24 yellow hard hats remain unchanged. \hfill \texttt{success\_rate=1.0}

\end{itemize}

\medskip
\textbf{Step 4.}
\begin{itemize}\setlength\itemsep{2pt}
  \item[{\small[1]}] 19 pink, 3 green, and 24 yellow hard hats remain in the truck. \hfill \texttt{success\_rate=0.0}
  \item[{\small[1]}] 16 pink, 3 green, and 24 yellow hard hats remain in the truck. \hfill \texttt{success\_rate=0.75}
  \item[{\small[2]}] 16 pink and 3 green hard hats remain; $16+3=19$. \hfill \texttt{success\_rate=0.875}

  \item[{\small[3]}] Total remaining hats: $16+3+24=43$. \hfill \texttt{success\_rate=0.875}
\item[{\small[4]}] 16 pink and 3 green hard hats remain, so $16+3=19$ hats across these two colors. \hfill \texttt{success\_rate=1.0}

\end{itemize}

\medskip
\textbf{Step 5.}
\begin{itemize}\setlength\itemsep{2pt}
  \item[{\small[0]}] Total: $3+19+24=?$ \hfill \texttt{success\_rate=0.0}
   \item[{\small[1]}] Total remaining hats: $3+19+24=?$ \hfill \texttt{success\_rate=0.0}
  \item[{\small[2]}] Total: $19+24=43$. \hfill \texttt{success\_rate=0.875}
  \item[{\small[3]}] Combining 19 (pink+green) with 24 yellow gives $19+24=43$. \hfill \texttt{success\_rate=1.0}
  \item[{\small[4]}] Total remaining hats: $19+24=43$. \hfill \texttt{success\_rate=1.0}

\end{itemize}

\end{tcolorbox}
\caption{Representative dataset example with step-wise reasoning annotations.}
\label{fig:dataset_example}
\end{figure*}



\end{document}